\documentclass[10pt,twocolumn,letterpaper]{article}

\usepackage{cvpr}              
\usepackage{caption}
\usepackage{graphicx}
\usepackage{multirow}
\usepackage{amssymb}
\usepackage{bbding}
\usepackage{makecell}
\usepackage{pifont}
\usepackage[accsupp]{axessibility} 
\usepackage[dvipsnames]{xcolor}

\definecolor{cvprblue}{rgb}{0.21,0.49,0.74}
\usepackage[pagebackref,breaklinks,colorlinks,citecolor=cvprblue]{hyperref}

\def\paperID{5052} 
\def\confName{CVPR}
\def\confYear{2024}

\title{Joint2Human: High-quality 3D Human Generation via\\ Compact Spherical Embedding of 3D Joints}

\author{
    Muxin Zhang$^{1, \dagger}$, Qiao Feng$^{1, \dagger}$, Zhuo Su$^{2}$, Chao Wen $^{2}$, Zhou Xue$^{3}$, Kun Li$^{1,*}$ \\
    $^{1}$College of Intelligence and Computing, Tianjin University \\ $^{2}$PICO IDL, ByteDance China $^{3}$Li Auto  \\
{\tt\small }
}

\begin{document}
\twocolumn[{
\maketitle
\begin{center}
    \captionsetup{type=figure}
    \includegraphics[width=1.0\textwidth]{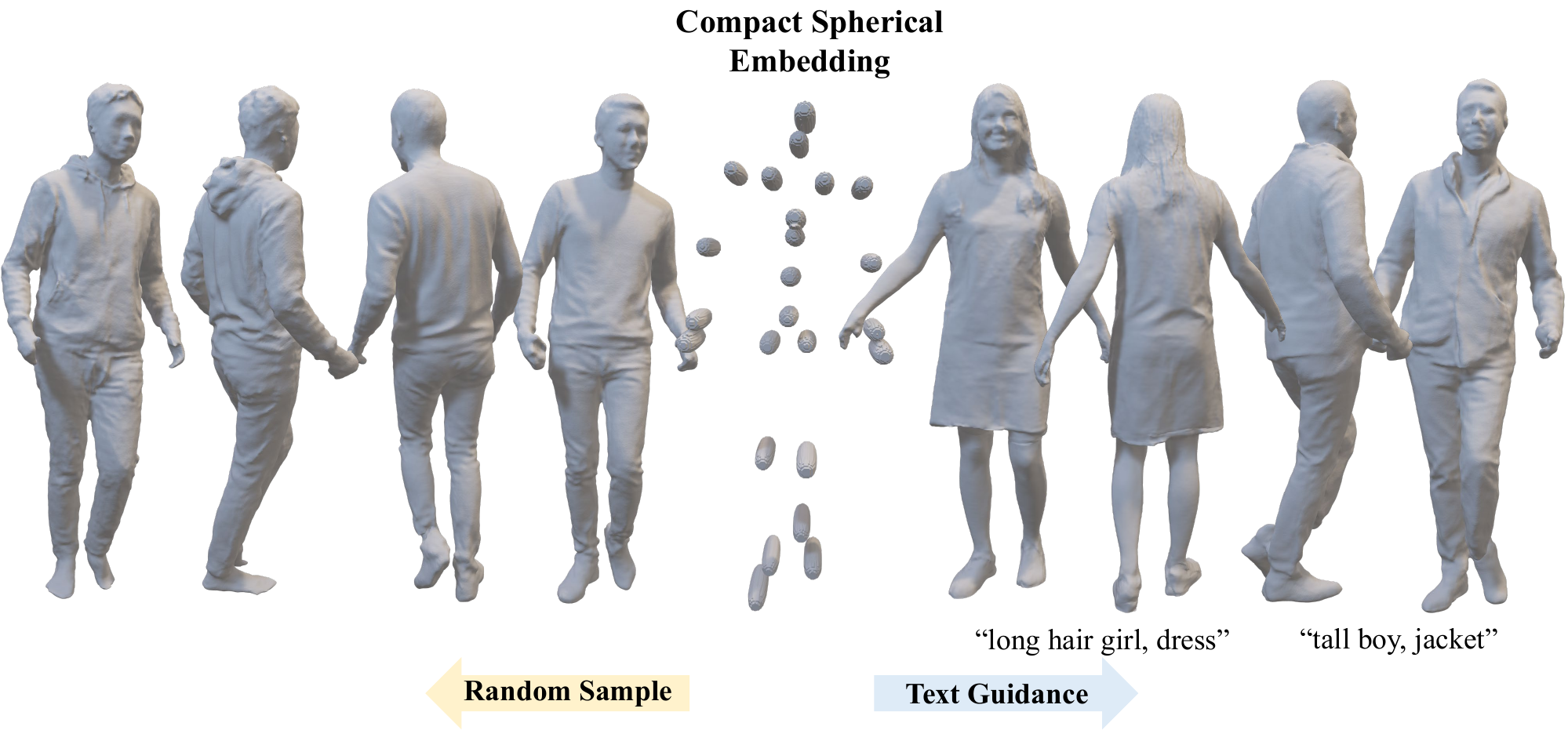}
    \vspace{-0.95cm} 
    \captionof{figure}{With the proposed compact spherical embedding of 3D joints, our method can generate high-quality and high-resolution 3D humans with reasonable global structure and fine-grained geometry details in an efficient way, based on a native conditional 3D generative network with 2D diffusion model.}
    \label{fig1}
\end{center}
}]
\let\thefootnote\relax\footnotetext{$\dagger$ Equal contribution.}

\let\thefootnote\relax\footnotetext{* Corresponding author.}
\begin{abstract}
3D human generation is increasingly significant in various applications. However, the direct use of 2D generative methods in 3D generation often results in losing local details, while methods that reconstruct geometry from generated images struggle with global view consistency. In this work, we introduce Joint2Human, a novel method that leverages 2D diffusion models to generate detailed 3D human geometry directly, ensuring both global structure and local details. To achieve this, we employ the Fourier occupancy field (FOF) representation, enabling the direct generation of 3D shapes as preliminary results with 2D generative models. With the proposed high-frequency enhancer and the multi-view recarving strategy, our method can seamlessly integrate the details from different views into a uniform global shape. To better utilize the 3D human prior and enhance control over the generated geometry, we introduce a compact spherical embedding of 3D joints. This allows for an effective guidance of pose during the generation process. Additionally, our method can generate 3D humans guided by textual inputs. Our experimental results demonstrate the capability of our method to ensure global structure, local details, high resolution, and low computational cost simultaneously. More results and the code can be found on our project page at \url{http://cic.tju.edu.cn/faculty/likun/projects/Joint2Human}.

\end{abstract}    
\section{Introduction}
\label{sec:intro}

The field of 3D human generation holds considerable significance across diverse applications, including virtual/augmented reality, gaming, and the film industry.
Despite advancements in existing methods~\cite{kim2023chupa, dong2023ag3d, xiong2023Get3DHuman}, it remains a challenge to achieve a simultaneous guarantee of both global structural accuracy and local geometry details, which is often compounded by high computational costs.
In this paper, we aim to address these limitations and propose a method for generating high-quality 3D human models that exhibit fidelity to both global structures and local details while ensuring computational efficiency, as illustrated in Figure \ref{fig1}. 

%
Existing human generation methods can be divided into two categories: two-stage methods and native 3D generation methods. Two-stage generation methods~\cite{hong2023evad,jiang2022humangen,xiong2023Get3DHuman,kim2023chupa,orel2022stylesdf} fit 3D humans from 2D images via NeRF~\cite{mildenhall2020nerf} or differentiable rendering~\cite{DVR}.
These methods are trained only on 2D datasets and hence lack a 3D backbone for perceiving 3D structures and ensuring view consistency.
Chupa~\cite{kim2023chupa} aims to produce high-quality results through dual normal map-based optimization but suffers from depth ambiguity issues that harm view consistency.
Most native 3D generation methods~\cite{bergman2022gnarf,ENARF-GAN,dong2023ag3d,wang2023rodin} generate 3D humans directly with Tri-planes~\cite{Chan2021EG3D} or Signed Distance Fields~\cite{orel2022stylesdf}. 
%
However, these approaches encounter challenges in achieving high-fidelity global structures with fine-grained local geometry details. 
Additionally, their generalization capabilities are limited, and computational efficiency is compromised.
In summary, the existing methodologies collectively fail to ensure global structural fidelity, local detail preservation, high resolution, and computational efficiency simultaneously.

To efficiently generate high-quality 3D humans with reasonable global structure and fine-grained geometry details, in this paper, we propose \textit{Joint2Human}, a conditional generative network with 2D diffusion models derived from 3D datasets.
%
To achieve high-fidelity 3D human generation with reasonable global structure, we employ the 2D generative models to produce 3D shapes as preliminary results directly. Subsequently, we carefully design the compact spherical embedding of 3D joints based on an image-aligned 3D representation FOF~\cite{li2022neurips,feng2022monocular}. With it, we implement effective pose guidance and diverse generation.
We also design a high-frequency enhancer and a multi-view recarving strategy for fine-grained local detail generation.
Experimental results demonstrate that our method outperforms the state-of-the-art methods regarding global structure, local detail, and computational efficiency. 
Furthermore, our method also exhibits versatility by enabling the generation of 3D human representations guided by text. 
 
To summarize, our main contributions include:
\begin{itemize}
    \item We propose \textit{Joint2Human}, a native conditional 3D generative method with a 2D diffusion model for high-quality, high-resolution 3D human generation. To our knowledge, it is the first work based on the FOF that can simultaneously ensure global structure, local details, high resolution, and low computational cost.
    \item We propose a new pose guidance embedding, a compact spherical embedding of 3D human joints, for efficient perception of global structure. This mechanism also facilitates a more straightforward and effective implementation of pose-guided generation in 2D generation framework.
    \item We design a high-frequency enhancer by integrating a subsidiary decoder into the pre-trained VAE and a multi-view recarving strategy for fine-grained local detail generation. Both of them improve the geometry quality of the final results.
\end{itemize}

\section{Related Work}
\label{sec:formatting}
\subsection{3D Human Generation with 2D Generators}
Many approaches~\cite{hong2023evad,jiang2022humangen,xiong2023Get3DHuman,kim2023chupa,orel2022stylesdf,gao2022get3d} try to learn the 3D shape from 2D images via various NeRF representations~\cite{mildenhall2020nerf,2021narf,liu2021neural,peng2021neural,peng2023implicit,kwon2021neural} and differentiable volume rendering~\cite{DVR,Zhao_2022_CVPR,weng_humannerf_2022_cvpr}. However, it is always computationally expensive and limited in resolution. EVA3d~\cite{hong2023evad} designs complex training strategies to achieve high-resolution generation, but its geometric quality is mediocre. What's worse, the 2D dataset is always imbalanced in viewing angles and human poses. Hence, these methods are relatively unstable, and it is hard to generate a realistic full-body human geometry. Other methods ~\cite{jiang2022humangen,xiong2023Get3DHuman,kim2023chupa} also leverage the priors from 2D human generation and 3D human reconstruction models, which is cumbersome. In Chupa~\cite{kim2023chupa}, many optimizations are needed to maintain view consistency. It first generates two normal maps for a clothed human's front and back sides, then optimizes the 3D generated mesh by the dual normal maps. There are also some human template-based works ~\cite{liao2024tada,zhang2022avatargen,hong2022avatarclip,youwang2022clipactor,cao2023dreamavatar,kim2023chupa} that are highly dependent on human template mesh, such as SMPL~\cite{SMPL:2015} and SMPL-X~\cite{SMPL-X:2019}, which adversely affect the diversity of generated models especially in local details. For example, it is flawed in modeling loose clothes like dresses.

\begin{table*}[!t]
  \small
  \caption{Comparison with existing methods. }
  \label{table1}
  \centering

  \resizebox{0.95\textwidth}{!}{
  \begin{tabular}{lccccc}
    \toprule
    
    Method & \makecell[c]{Global Structure\\ (View Consistency)} &

    \makecell[c]{Local Detail}&
    \makecell[c]{High Resolution} &
    \makecell[c]{Loose Clothes} &
    \makecell[c]{Computational\\Efficiency }\\
    \midrule
     GNARF\cite{bergman2022gnarf} &\Checkmark & \XSolidBrush & \XSolidBrush & \XSolidBrush &\XSolidBrush \\   
    AG3D \cite{dong2023ag3d}  & \Checkmark & \XSolidBrush &  \XSolidBrush  &  \XSolidBrush &\XSolidBrush\\
    Get3Dhuman \cite{xiong2023Get3DHuman}   & \XSolidBrush & \Checkmark & \Checkmark & \Checkmark &\XSolidBrush\\
    Chupa \cite{kim2023chupa}  & \XSolidBrush & \Checkmark &\Checkmark &  \XSolidBrush &\XSolidBrush \\
    \textbf{Ours} & \Checkmark & \Checkmark & \Checkmark & \Checkmark &\Checkmark\\
    \bottomrule
  \end{tabular}
  }
\end{table*}

\subsection{3D-aware Generation}
 Based on the development of 3D human representation, numerous methods have emerged for native 3D generation. Some of these approaches~\cite{bergman2022gnarf,ENARF-GAN,dong2023ag3d,wang2023rodin} utilize tri-planes~\cite{Chan2021EG3D} which is a NeRF-based representation. It is hard to generate precise geometry, which is limited by computational complexity. Apart from that, the implicit functions ~\cite{orel2022stylesdf,park2019deepsdf} are highly favored in several methods~\cite{xiong2023Get3DHuman,cao2023dreamavatar,hong2023evad,hong2022avatarclip,kolotouros2023dreamhuman}. Alternatively, some methods use 3D data for training, gDNA~\cite{chen2022gdna} use implicit multi-subject forward skinning which enables learning from 3D scans of human bodies. Among all the works mentioned above, the GAN~\cite{goodfellow2014generative,zhang2022avatargen,hong2023evad,bergman2022gnarf,ENARF-GAN,dong2023ag3d,jiang2022humangen,Chan2021EG3D,orel2022stylesdf,xiong2023Get3DHuman} and Diffusion models~\cite{ho2020denoising,wang2023rodin,cao2023dreamavatar,kim2023chupa,kolotouros2023dreamhuman} are the most popular. 

Using the diffusion model for small object generation\cite{cheng2023sdfusion,li2023diffusionsdf,chou2022diffusionsdf,shapenet} is more straightforward to combine with text control. However, the text-guided 3D human generation is more difficult due to the high-dimensional cube and lack of text-3D data pairs. Some attempts were made to tackle these issues. AvatarCLIP~\cite{hong2022avatarclip} initialize a bare mesh shape with SMPL. It continuously optimizes the mesh to match the input text by calculating the CLIP score between the text token and the mesh's rendering image. CLIP-actor~\cite{youwang2022clipactor} does the same thing, except it designs a recommendation module to get the initial shape/motion. In this way, they reduced the need for text-3D data pairs with the help of the large text-image pre-trained model~\cite{radford2021learning}. Other methods~\cite{kim2023chupa,cao2023dreamavatar,wang2023rodin} impose textual controls on the diffusion model for generated shapes that match the input description. In previous approaches~\cite{chou2022diffusionsdf,li2023diffusionsdf,cheng2023sdfusion} to generating small objects, the resolution of the generated geometry is limited for human generation, such as $64^{3}$ or $128^{3}$. In addition, some methods, such as differentiable rendering-based methods, are time-consuming to infer.

Different from the two types of methods described above, as shown in \cref{table1}. Our approach can accommodate both global structures and local details. To fully perceive the global structures, we adopt the image-aligned 3D representation FOF and introduce a compact spherical embedding of 3D joints for pose guidance. To generate details, we design a high-frequency enhancer and a multi-view recarving strategy in 3D space. Apart from this, we also simultaneously achieve high-resolution generation and low computational cost.

\section{Method}
\begin{figure*}[t]
  \centering
   \includegraphics[width=1.0\textwidth]{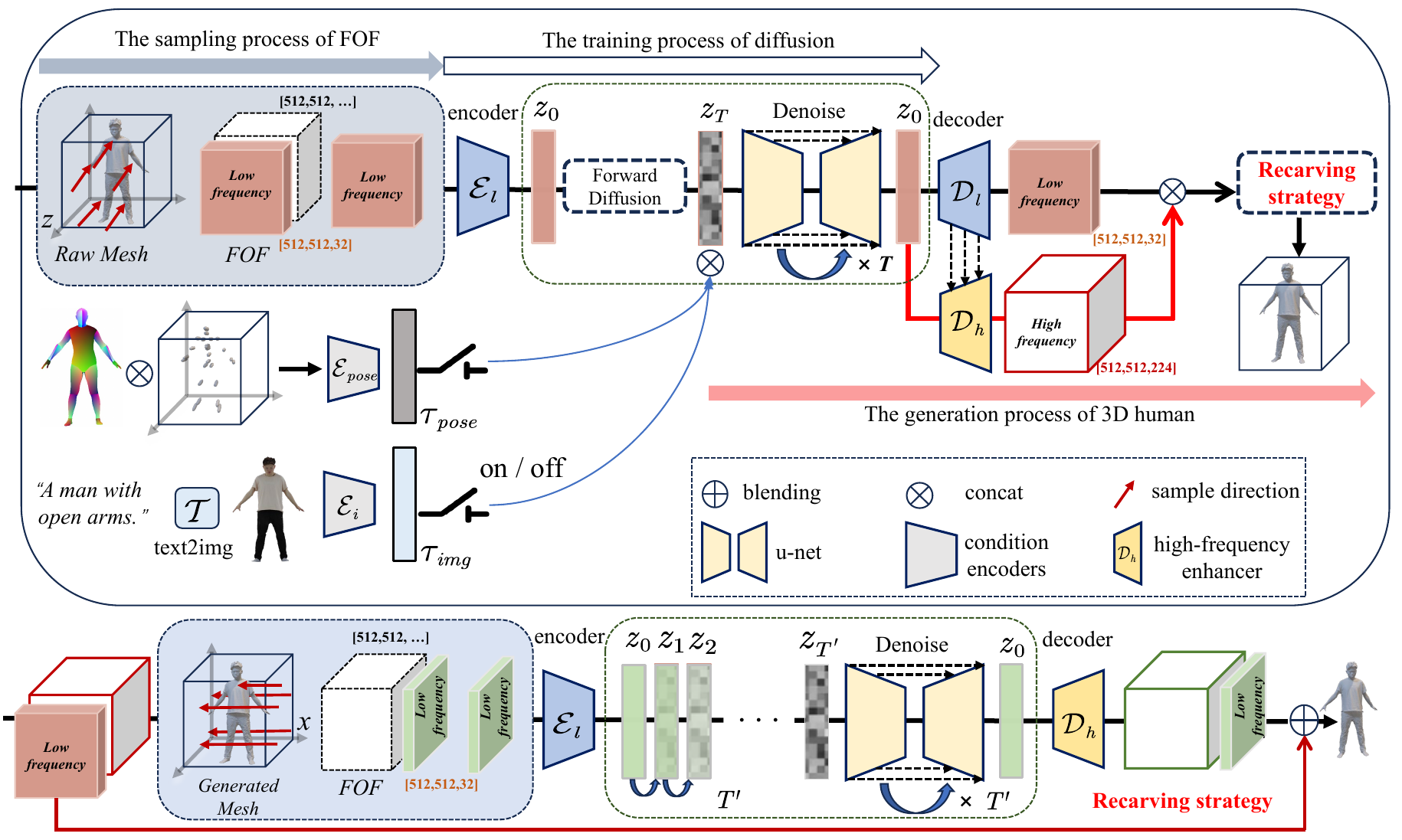}
   \caption{\textbf{Overview of our framework.} \textit{Joint2Human} takes Gaussian noise and some conditional information as input. We first encode FOF into a latent space, where a diffusion model is trained to enable high-resolution human generation. Furthermore, to enable flexible conditional generation, we adopt condition encoders along with classifier-free guidance to enable conditional generation. Our conditional control generation strategy can support switching between different modalities. Then, we pass the results generated by the first denoising process through the high-frequency enhancer $\mathcal{D}_h$ and multi-view recarving strategy for fine-grained local detail generation. Additionally, the recaiving strategy is shown in the bottom subfigure. }
    \label{fig:pipeline}
\end{figure*}

\begin{figure}[t]
  \centering
   \includegraphics[width=1.0\linewidth]{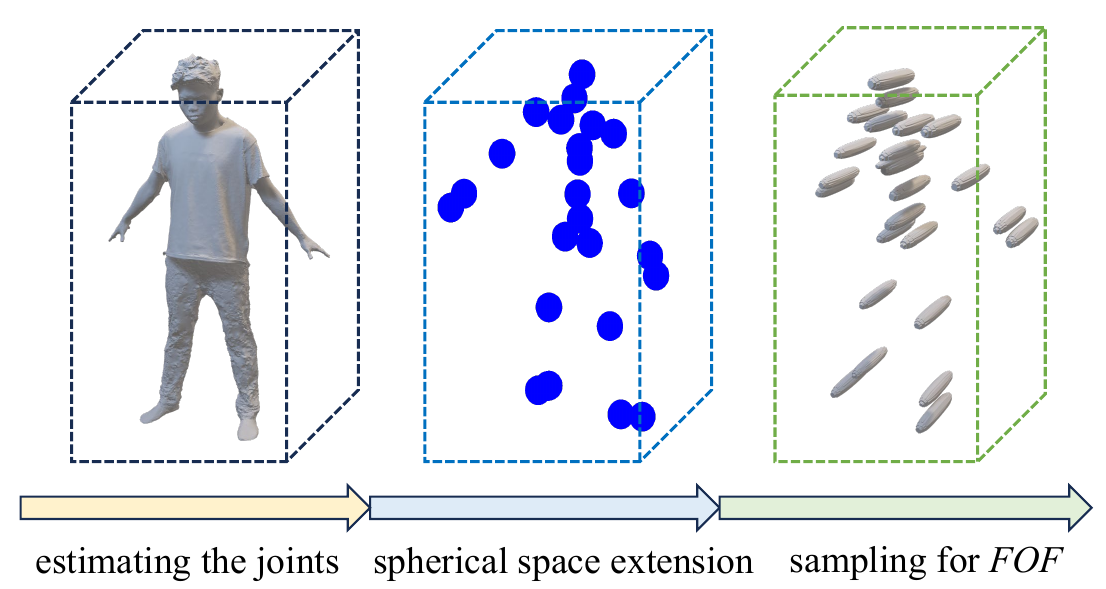}
   \caption{The data processing flow for compact spherical embedding of 3D joints.}
   \label{fig:joint}
\end{figure}

Our method aims to generate diverse 3D-clothed humans with global structure and local details. First, we propose a direct conditional 3D generative method with 2D diffusion models as illustrated in \cref{fig:pipeline}. Specifically, we utilize an image-aligned 3D representation FOF to produce 3D shapes directly. We operate the condition and generation process in the latent space (\cref{sec:latent}). Then, we propose the compact spherical embedding of 3D joints (\cref{sec:condition}) for precise pose control and diverse generation. Furthermore, a high-frequency enhancer (\cref{sec:hfe}) and a multi-view recarving strategy (\cref{sec:mvrs}) are proposed to improve the geometry details further. We will detail all the stages of our method in the following subsections.

\subsection{Latent Diffusion for Fourier Occupancy Fields}
\label{sec:latent}
To model the distribution of Fourier Occupancy Fields, following the latent diffusion model~\cite{rombach2021highresolution}, we utilize an auto-encoder to compress the high-dimensional raw data space into a lower-dimensional latent space, which encodes human shapes into a normal distribution. In detail, we adopted a VAE~\cite{kingma2013auto}, which contains the encoder $\mathcal{E}  $ and decoder $\mathcal{D}$. Given a FOF feature $x\in \,\,\mathbb{R} ^{512\times 512\times 32}$, the encoder $\mathcal{E}$ encodes FOF into latent vectors $z=\mathcal{E} \left( x \right)$, where $ z \in \mathbb{R} ^{128\times 128\times 8}$. The decoder $\mathcal{D}$ decodes the latent vectors back to the original FOF space $\tilde{x}=\mathcal{D} \left( \mathcal{E} \left( x \right) \right)$, where $\tilde{x}\in \mathbb{R} ^{512\times 512\times 32}$. We pretrained the auto-encoder with the reconstruction loss and KL-regularization loss like~\cite{rombach2021highresolution}. 
 During training, the training data is passed through the encoder $\mathcal{E}$ in the vae module to obtain its feature in the latent space. We adopt a U-Net-like structure as our denoising model $\epsilon_{\theta}$. We follow the classical and efficient loss function proposed by Ho~\cite{ho2020denoising} and train the diffusion model based on the $T$ steps of noise-adding and denoising processes in the latent space. Thus, we can obtain a vector $\tilde{z}_0$ in the latent space, which will be sent to the decoder $\mathcal{D}$ to get FOF. Then, we can convert the FOF to an occupancy field and get an initial 3D human mesh from it with the Marching Cubes~\cite{lorensen1998marching} algorithm.

\subsection{Condition-guided Generation Mechanisms}
\label{sec:condition}
\noindent \textbf{Compact Spherical Embedding of 3D Joints.} 
Different from previous works~\cite{kim2023chupa,dong2023ag3d,zhang2022avatargen,youwang2022clipactor,hong2022avatarclip,cao2023dreamavatar} adopt the human parametric model~\cite{SMPL:2015,SMPL-X:2019} for pose guidance, we find that such a strategy with FOF can lead to overfitting. 
Conditioning on the SMPL makes it harder for networks to learn the true data distribution, which is also mentioned in ECON~\cite{xiu2023econ}. To avoid the redundant information and the misleading geometric prior from SMPL, we use the human body joints as pose guidance and design a compact spherical embedding of 3D joints. As shown in \cref{fig:joint}, we first estimate the $K$ 3D joints location  $J=\left\{ p_i\in \mathbb{R} ^3 \right\} _{i=1}^{K}$ of the given human. For each joint position $p_i$, we extend a sphere with $p_i$ as the center and $r$ as the radius in 3D space. After that, according to the \cref{sec:latent}, we compute the FOF $c_i\in \mathbb{R} ^{512\times 512\times 8}$ for each sphere. Therefore, the depth information is stored in the phase of trigonometric functions. We concatenate $\left\{ c_i \right\} _{i=1}^{K}$ channel-wise with a fixed order to form the compact spherical embedding of 3D joints $con_{joint}\in \mathbb{R} ^{512\times 512\times 8K}$. In this way, our approach processes joints in different channels and
concatenates them to form the embedding of 3D joints, integrating full semantic and depth-wise information for precise pose control. In our setting, we use $r = 10 cm$ and follow the joints $K = 24$ in SMPL. \\

\noindent \textbf{Pose-guided Geometry Generation.} We aim to generate a human shape that fits the given pose. With the compact spherical embedding of the 3D joints, we can efficiently perform pose control in human generation. To further improve the stability of the generative model, we incorporate the 2D prior by utilizing the IUV map $con_{iuv}$ defined in DensePose~\cite{guler2018densepose}. 
During training, we first get the $con_{pose}$ by concatenating the $con_{joint}$ and $con_{iuv}$ in the channel dimension. Then, we learn a pose encoder $\varepsilon _{pose}$ to get the human pose embedding $\tau_{pose}$ and apply the pose condition $con_{pose}$ on the latent code $z$ by concatenation after passing through the conditional encoder :
\begin{equation}
    \label{pose_encoder}
    \tau _{pose}=\varepsilon _{pose}\left( con_{pose} \right) ,
\end{equation}
where the $\varepsilon _{pose}$ consists of two convolutional layers for the alignment of the feature.
In the inference stage, we first get the human pose embedding $\tau_{pose}$; then we perform the denoising process of the $\epsilon_{\theta}$ to get the initial human shape.  \\

\noindent \textbf{Text-guided Geometry Generation.} 
 To generate the 3D human with the guidance of text description, we proposed the image prompt strategy for our model based on a text-guided 2D human image generation model Text2Human~\cite{jiang2022text2human}. In the training stage, we learn an additional image encoder $\varepsilon _{i}$ to get the human image embedding $\tau_{img}$, then apply it to the training of the diffusion model by concatenation. In the inference stage, with the input text, we first infer the text-to-image model $\mathcal{T} $ and image encoder $\varepsilon _{i}$ to get the embedding $\tau_{img}$, then we sample the 3D human shape from $\epsilon_{\theta}$ with all these conditions. We provide additional text-guided generation results in the supplementary materials.

\subsection{High-frequency Enhancer}
\label{sec:hfe}
For efficient human representation and modeling of data distribution, we only save low-frequency terms while taking the FOF as a 3D representation in ~\cref{sec:latent}. However, some abandoned high-frequency information is essential for human shape details. Therefore, based on the generation of the low-frequency FOF feature, we propose the High-Frequency enhancer to recover these missing details $C_h$, as shown in \cref{fig:pipeline}. In detail, we learn a reference-based decoding network $\mathcal{D}_h$ based on the latent space and decoder from the auto-decoder claimed in \cref{sec:latent} to predict high-frequency terms $\hat{C}_h$: 
\begin{equation}
    \label{l2h}
     \hat{C}_h=\mathcal{D} _h\left( \mathcal{D} _l\left( z_0 \right) ,z_0 \right) ,
\end{equation}
To enhance the perception of latent-space contextual information, we introduce additional skip connections for each layer in $\mathcal{D}_h$, which is used to fuse the feature map $\mathcal{D} _l\left( z_0 \right)$ from the relevant layers in the decoder $\mathcal{D}_l$. When training, we calculate the MSE loss between the predicted terms ${C}_h$ and high-frequency ground truth ${C}_h$ as the supervision to optimize $\mathcal{D}_h$,the loss function $\mathcal{L} _h$ is formulated as:
\begin{equation}
    \label{enhancer_loss}
    \mathcal{L} _h=\frac{1}{n}\sum_{i=1}^n{\left( \hat{C}_h\left( x_i,y_i \right) -C_h\left( x_i,y_i \right) \right) ^2} ,
\end{equation}
where $n$ is the number of all pixels in ${C}_h$. So, in the inference stage, we can estimate the high-frequency feature based on the known low-frequency feature to enhance the geometric details.

\subsection{Multi-view Recarving Strategy}
\label{sec:mvrs}
Since our training data is sampled along a fixed direction and the directionality of FOF, there are a few artifacts along the direction orthogonal to the normal direction of the generated shape. We propose the multi-view recarving strategy to improve the geometry quality while maintaining muti-view consistency.

Different from the resampling in Chupa~\cite{kim2023chupa}, we perform this process on 3D space and fuse them by blending the occupancy fields. 
In detail, we first perform the inference process of diffusion and high-frequency enhancer to obtain the FOF $C_{init}$ and the occupancy field $F_{init}$. Based on the $F_{init}$, we can get the initial human mesh $\mathcal{M}_{init}$, which may have some artifacts in other views. To tackle these artifacts, we rotate the $\mathcal{M}_{init}$ $\theta$ degrees along the yaw axis to get $\mathcal{M} _{\theta}$. After that, we convert the $\mathcal{M} _{\theta}$ into FOF $C_{\theta }$ along the current orientation. With the $C_{\theta }$, we do the same thing as described in  \cref{sec:latent}: leveraging the encoder $\mathcal{E}$ to get the latent code $\hat{z}$. Then, we re-execute the forward process of diffusion, adding $T^{\prime}$ steps of noise to $\hat{z}$ and denoising the noise $\hat{z}$ with $\epsilon_{\theta}$. We pass the denoised latent code through the decoder to get the FOF $C_{\theta}$. Meanwhile, we can reconstruct the occupancy field $F_{\theta}$ from FOF $C_{\theta}$ by efficient Fourier inversion. At last, we blend the occupancy fields $\left\{ F_{init},F_{\theta _1},F_{\theta _2}... \right\} $ of different views and extract the 3D human mesh from the occupancy fields. In our setting, we perform this procedure only once with the $\theta =\frac{\pi}{2}$. In this way, two orthogonal views are blended in a weighted-average approach. We settle on this setup under the trade-off between computational efficiency and geometric quality.

\section{Experiments}
\subsection{Experimental Setup}
\begin{itemize}
    \item \textbf{Datasets.} 
        We train our model with THuman 2.0~\cite{tao2021function4d}, THuman 3.0~\cite{deepcloth_su2022},  2k2k~\cite{han2023Recon2K} and about 1500 high-quality meshes from commercial datasets. To ensure a fair comparison, we deviated from Chupa~\cite{kim2023chupa}'s settings and used a third-party dataset CustomHumans~\cite{ho2023learning} as our test set. This dataset wasn't used to train any model. We do this because the Sota methods’ training code is not publicly available, and some of the datasets they use are not publicly available. For the training stage, we first sample 32 successive different angles of FOF along the yaw axis of the rotating human body to generate FOF feature maps for the same mesh. After that, we obtain the joint points for each mesh and calculate the compact spherical embedding of the 3D joint using the same FOF sampling process.
    \item \textbf{Baselines.}
        We compare our method with Chupa~\cite{kim2023chupa} and AG3D~\cite{dong2023ag3d} as baselines. Chupa is the current state-of-the-art method for generating 3D human geometry. We don't make a quantitative comparison with AG3D because it uses 2D datasets, and the training code of AG3D is unavailable. 
    \item \textbf{Metrics.}
        We measure the quality of the generated human mesh by using the Fr\'echet Inception Distance (FID)~\cite{fid} between the rendering normal maps~\cite{chen2022gdna,kim2023chupa} and shading images~\cite{kim2023chupa,Shue20223DNF} of the generated meshes. We follow the settings of Chupa~\cite{kim2023chupa} and render the meshes into 18 views with $20^{\circ}$ yaw interval for calculating FID.
\end{itemize}

\subsection{Implementation Details}
 To represent and generate a high-quality 3D human geometry, the channels of the FOF feature maps need to be set to at least 32. For the training stage of the FOF auto-encoder, we train it for three days on 8 NVIDIA A100 GPUs, with a batch size of 32. For the diffusion model training stage, we train it for eight days on 8 NVIDIA A100 GPUs, with a batch size of 64. The total number of time steps is set as $T = 1000, T^\prime=200$ for the diffusion model in our pipeline. We present more details in the supplementary material. 
\begin{table}[!t]
\centering
\caption{Quantitative evaluation. We report two types of FID scores on the test dataset.}
\label{table_1}
\setlength{\tabcolsep}{14pt}
\begin{tabular}{lcc}
\toprule
Method & FID$_{normal} \downarrow$ & FID$_{shade}  \downarrow$\\
\hline
Chupa$_{coarse}$ & 51.60  \quad  & 73.03     \\
Chupa$_{fine}$   & 29.90  \quad  & 45.49     \\
Ours$_{coarse}$  & 37.67  \quad  & 59.45     \\
Ours$_{fine}$    &\textbf{23.89}  \quad  & \textbf{41.20}     \\
\bottomrule
\end{tabular}%
\end{table}

\begin{figure}[t]
  \centering
   \includegraphics[width=0.5\textwidth]{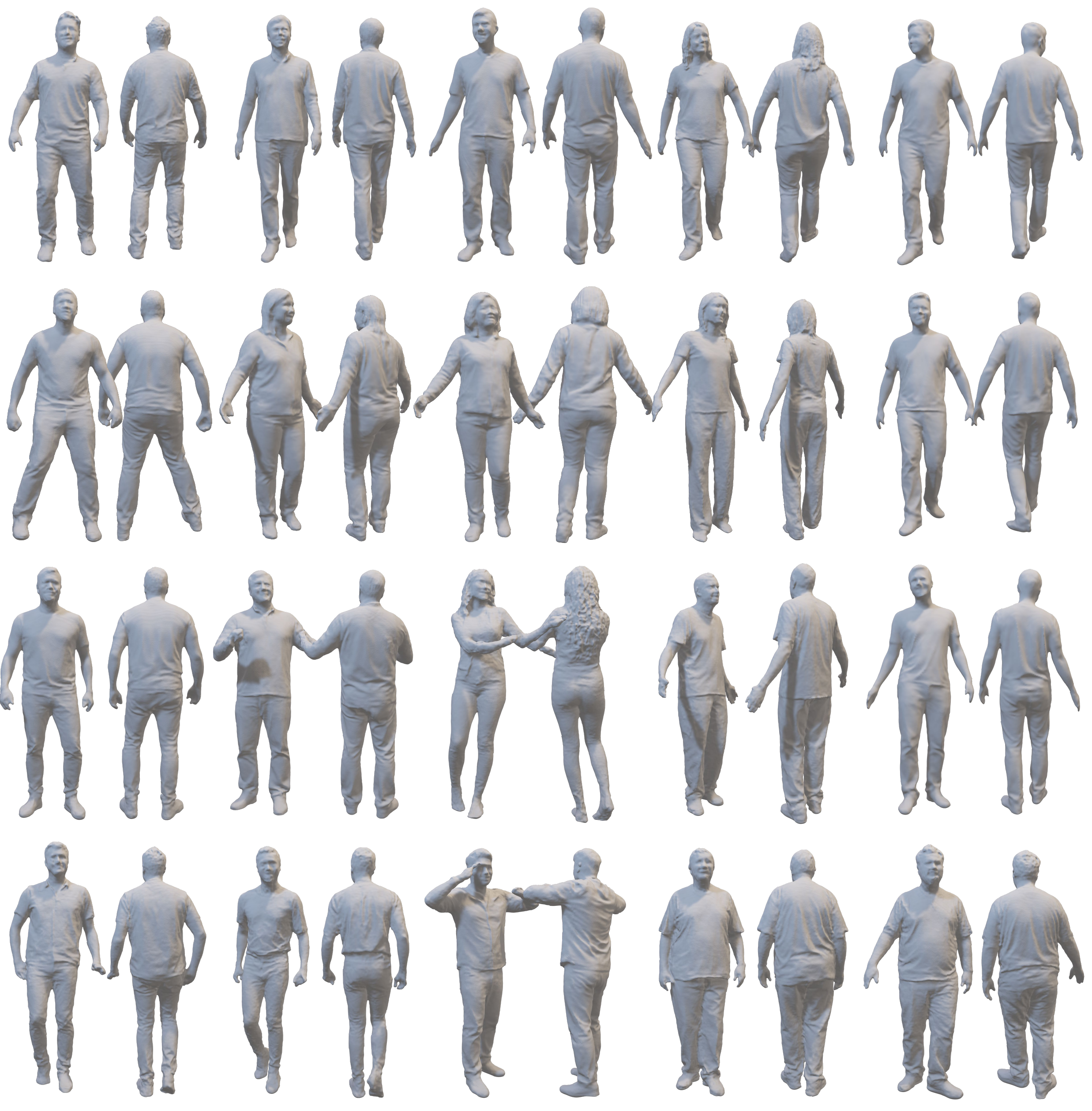}
   \caption{Our method can generate various high-quality 3D humans.}
    \label{fig:ours}
\end{figure}

\begin{figure}[t]
  \centering
   \includegraphics[width=0.48\textwidth]{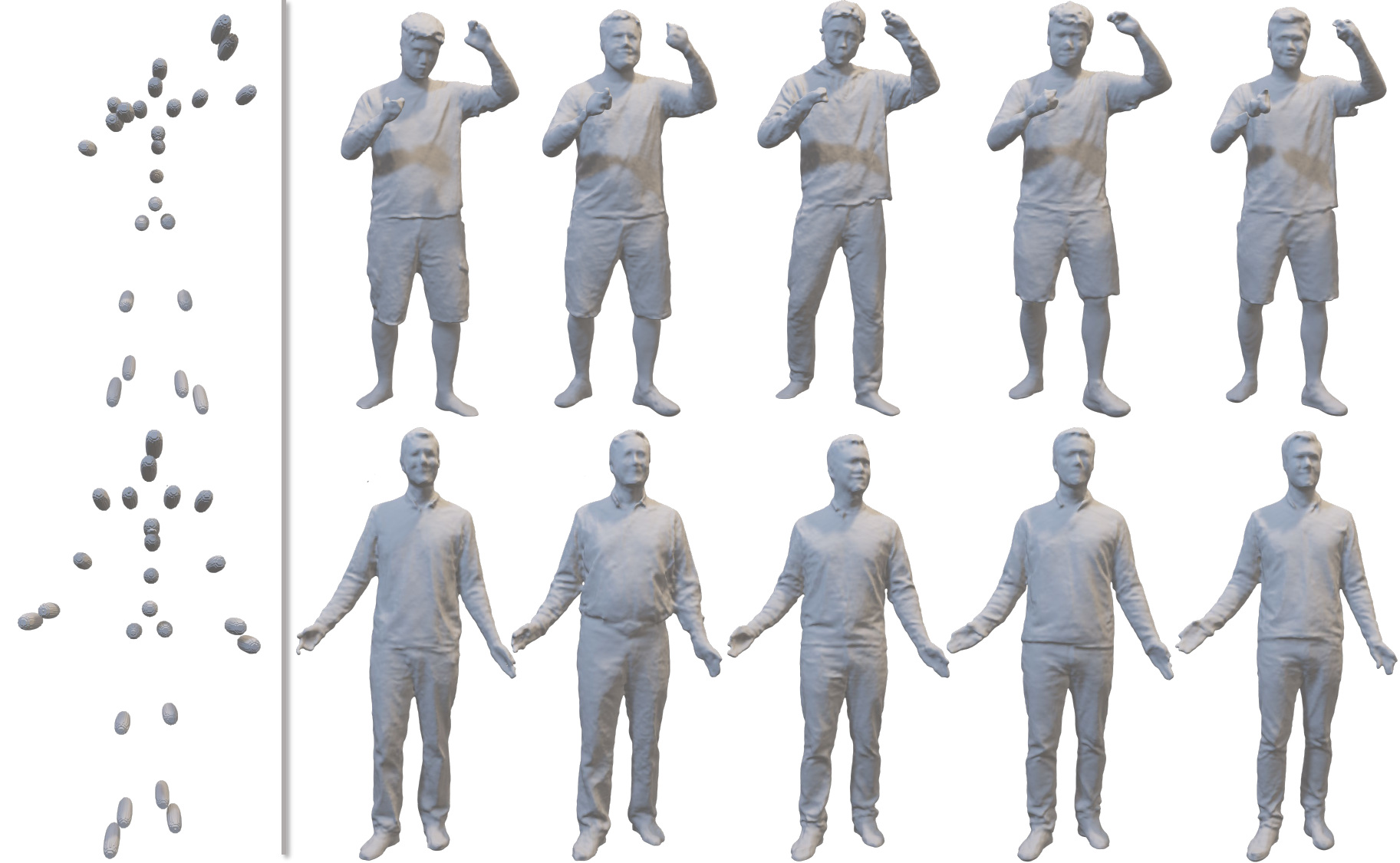}
   \caption{Our method can generate diverse results given a specific pose.}
    \label{fig:fig6}
    \vspace{-0.7cm}
\end{figure}


\begin{figure*}[t]
  \centering
   \includegraphics[width=1.0\textwidth]{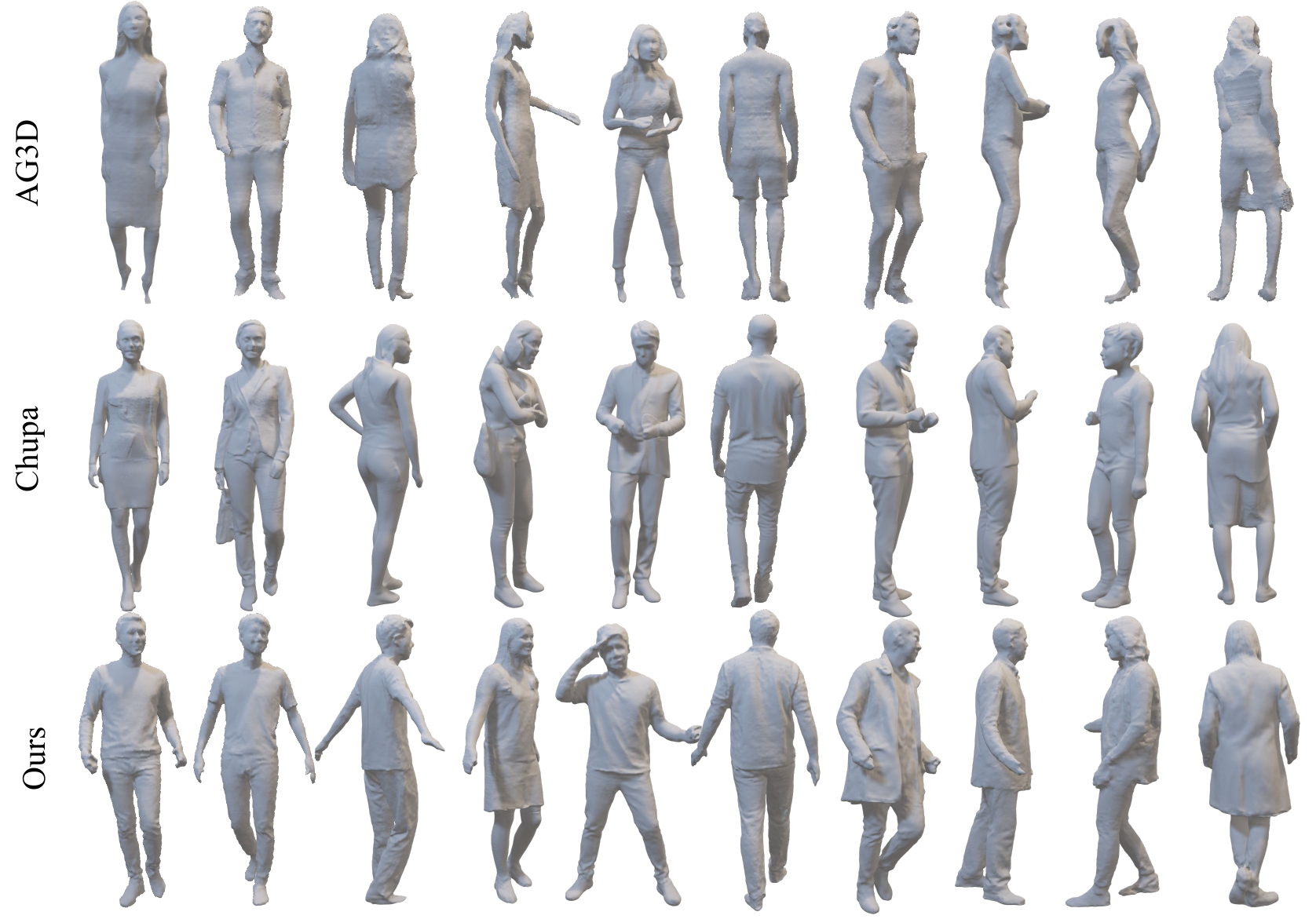}
   \caption{The generated 3D humans compared with Chupa~\cite{kim2023chupa} and AG3D~\cite{dong2023ag3d}.}
    \label{fig:compare}
\end{figure*}

 \begin{figure}[t]
  \centering
   \includegraphics[width=0.5\textwidth]{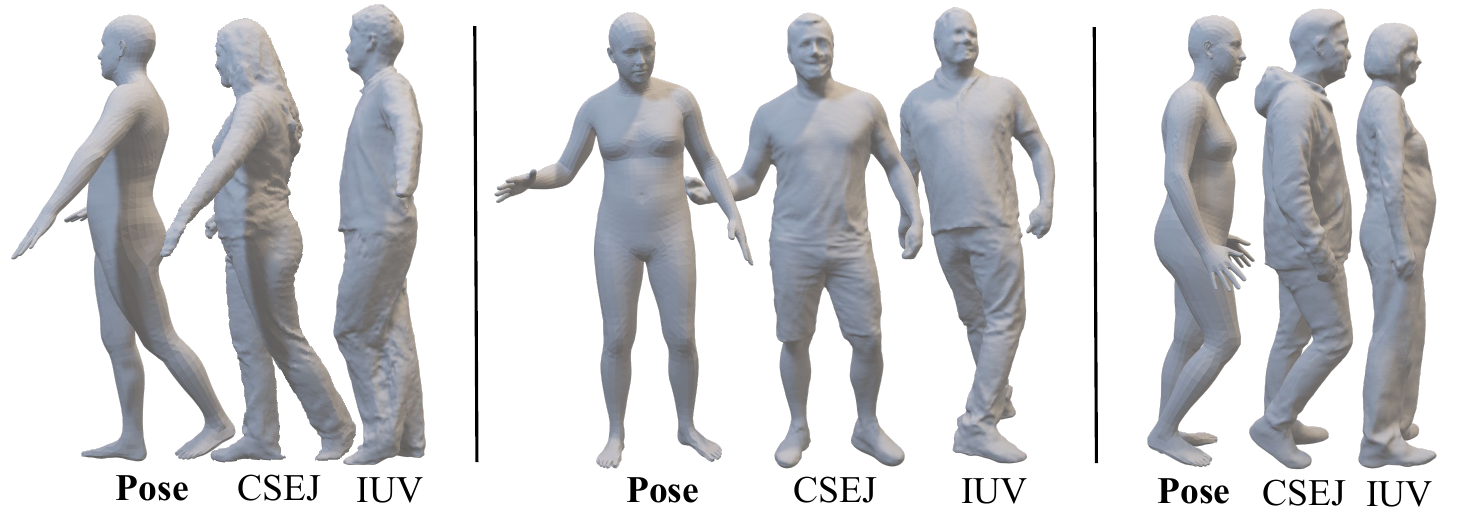}
   \caption{The generated results with different pose-guidance strategies.}
    \label{fig:pre_pose}
\end{figure}

\begin{figure}[t]
  \centering
   \includegraphics[width=0.45\textwidth]{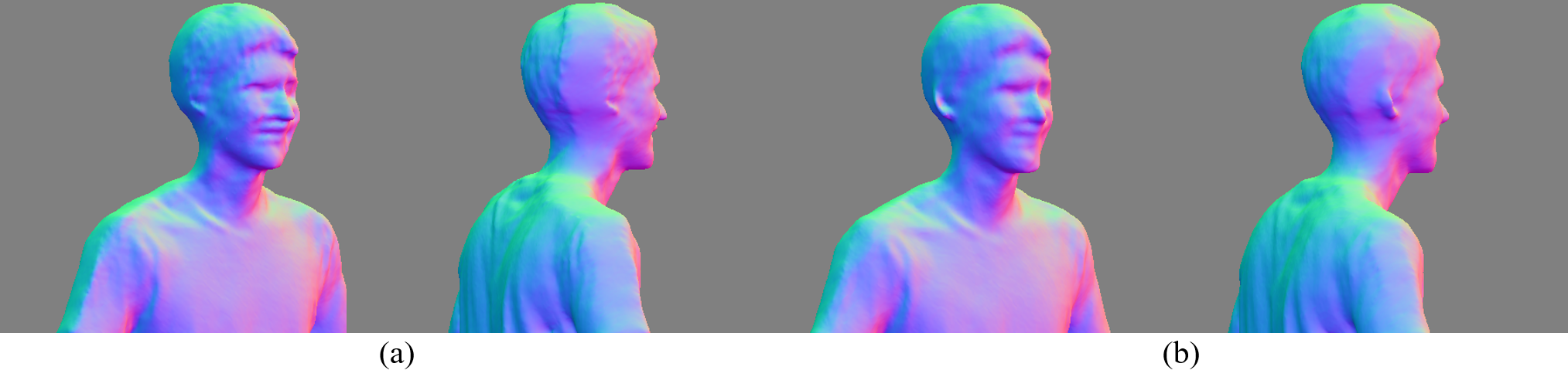}
   \caption{We present the side-view normal maps of the generated human. (a) illustrates the results without the High-Frequency Enhancer, while (b) illustrates the results using the High-Frequency Enhancer.}
    \label{fig:l2h_abl}
    \vspace{-0.5cm}
\end{figure}

 \begin{figure}[t]
  \centering
   \includegraphics[width=0.4\textwidth]{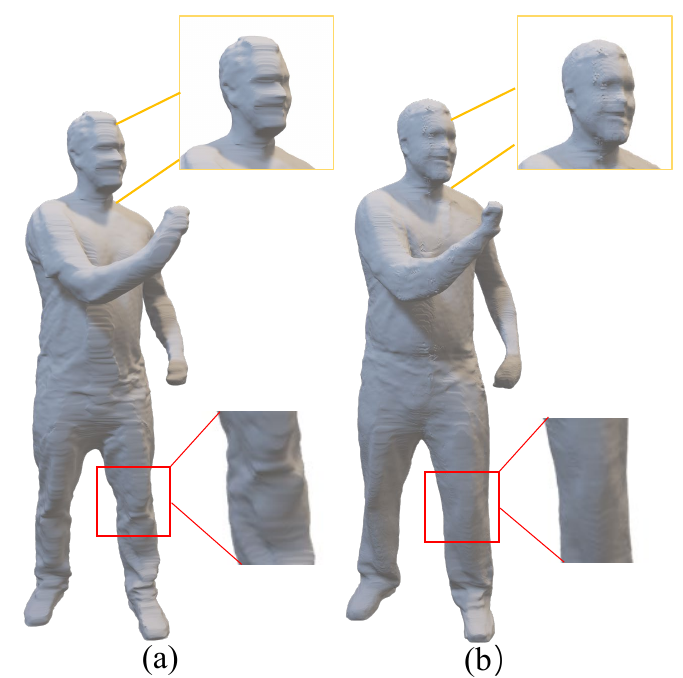}
   \caption{Ablation study on multi-view recarving strategy. With the multi-view recarving strategy (b), some artifacts can be avoided, compared with the variant without this strategy (a).}
    \label{fig:recarving}
    \vspace{-0.5cm}
\end{figure}

\subsection{Generated Results}
 \cref{fig:ours} shows various generated results by our method. Our method can generate high-quality 3D humans with global structure, local details, and high resolution. Benefiting from our proposed compact spherical embedding of 3D joints, we can flexibly generate high-quality 3D humans guided by poses as shown in \cref{fig:fig6}. Our model performs well in the diversity of generations with the fixed pose as guidance. It can generate humans with different identities and costumes.

\subsection{Comparison}
 \noindent \textbf{Qualitative and Quantitative Results.} We compare our method with the latest human generative models Chupa~\cite{kim2023chupa} and AG3D~\cite{dong2023ag3d}. In detail, We conduct a quantitative comparison with Chupa to assess the quality of the generated meshes. We also conduct a qualitative comparison with AG3D and Chupa. To ensure a fair comparison, we try to make the input poses of Chupa and our method the same as much as possible. However, AG3D is not friendly to fixed poses of the results, so we used random poses in the test stage. The quantitative results for human generation are shown in \cref{table_1}. Our method outperforms the current SOTA methods and achieves better FID on both images. \cref{fig:compare} shows qualitative comparison results. Our results gain the natural and detailed visual effects. Besides, we have better view consistency when compared to the generated results of other methods. In addition, our model can complete the generation process in less than one minute. It is advanced in loose clothing modeling and diversity generation. The discussion on running time and the visualized generation results are shown in the supplementary material. 

 \noindent \textbf{User Study.} To better evaluate our method with other state-of-the-art methods, we conduct a perceptual study to ask the users about their preferences of the three methods on the following three aspects: (1) Global structure, (2) Local detail, and (3) Overall impression. In the user study, we collected 123 answers from 54 females and 69 males of different ages (6 users under 18, 113 users between 18 and 40, 2 users between 40 and 60, and 2 users above 60). The results are shown in \cref{table_popularity}, demonstrating that our approach is the most popular throughout the user study. A total of 70.73\% of the users consider our methods' diversity superior to Chupa (Method B). More details about the user study can be found in the supplementary material.

\begin{table}[h]
  \small
  \caption{Proportion of popularity of different methods in different metrics.}
  \label{table_popularity}
  \setlength{\tabcolsep}{7pt}
  \centering

  \resizebox{0.5\textwidth}{!}{
  \begin{tabular}{lcccc}
    \toprule
     Method &AG3D (A) & Chupa (B) & Ours (C)\\ 
    \midrule
     Global Structure   &18.70\% &32.20\% &\textbf{49.10\%}\\
     Local Detail      &18.05\% &29.92\% &\textbf{52.03\%}\\
    Overall Impression    &19.02\% &29.27\% &\textbf{51.71\%}\\
    Diversity of Generation &-       &29.27\% &\textbf{70.73\%}\\
    \bottomrule
  \end{tabular}
  }
\end{table}

\subsection{Ablation Study}
We have conducted extensive ablation studies to validate the components and settings of our pipeline. We use the same pose guidance in quantitative comparisons.
We also show and analyze different qualitative results to demonstrate the effectiveness of our modules. 

\noindent \textbf{Ablation on Different Pose-guidance Strategies.} 
To validate the effectiveness of our pose-guidance, we train our model under different conditional information for pose-guidance. There are three types of conditional information in our study, SMPL~\cite{SMPL:2015}, CSEJ (our compact spherical embedding of 3D joints), and IUV~\cite{guler2018densepose}. We calculate the Fid scores under the different pose conditions. As shown in \cref{table_pose}, the CSEJ can generally improve the generation quality. While CSEJ alone is not as good as IUV on this metric, the advantage of our CSEJ is enabling precise pose control. The FID only calculates the distance between two distributions but does not measure the ability to control the pose. \cref{fig:pre_pose} shows generated results using different guidance under the fixed pose; our CSEJ alone achieves better pose-control results than IUV.\\

\begin{table}[h]
\vspace{-0.5cm}
\centering
\caption{Ablation study on different pose-guided strategy. We report FID scores for different combinations.}

\label{table_pose}
\setlength{\tabcolsep}{10pt}
\begin{tabular}{cccc}
\toprule
SMPL& CSEJ &IUV map & FID$_{normal} \downarrow$\\
\hline
\Checkmark & \XSolidBrush  & \XSolidBrush  &   49.16\\
\XSolidBrush  & \Checkmark & \XSolidBrush  &  45.73\\
\XSolidBrush  & \XSolidBrush  & \Checkmark &  39.51\\
\XSolidBrush  & \Checkmark & \Checkmark & \textbf{38.40}\\
\bottomrule
\end{tabular}%
\vspace{-0.5cm}
\end{table}

\noindent \textbf{Ablation on Multi-view Recarving Strategies and High-Frequency Enhancer.} \cref{fig:l2h_abl}  demonstrates the better visual effect of using High-Frequency Enhancer.
The visual comparisons presented in \cref{fig:recarving} demonstrate that the recarving strategy significantly enhances the local details of the generated human. To further measure the ability of other modules to capture local details, we conduct comparative experiments under different modules and calculate the FID scores. The results are shown in \cref{table_refine}.

\begin{table}[h]
\centering
\caption{Ablation study on multi-view recarving strategy and high-frequency enhancer. We report FID scores for different combinations.}
\label{table_refine}
\setlength{\tabcolsep}{10pt}
\begin{tabular}{ccc}
\toprule
Recarving Strategy&  Enhancer& FID$_{normal}  \downarrow$\\
\hline

\XSolidBrush  & \XSolidBrush  &  37.39\\
 \Checkmark& \XSolidBrush  &  31.90\\
\XSolidBrush  &\Checkmark &  29.51\\
\Checkmark& \Checkmark& \textbf{25.67}\\
\bottomrule
\end{tabular}%
\vspace{-0.5cm}
\end{table}

\section{Conclusion and Discussion}
\noindent \textbf{Conclusion.} In this paper, we introduce \textit{Joint2Human}, a novel and efficient method for directly generating detailed 3D human geometry using 2D diffusion models. We propose a compact spherical embedding of 3D joints for flexible control and the utilization of human prior. We also design a high-frequency enhancer and a multi-view recarving strategy to seamlessly integrate the details from different views into a uniform global shape, guaranteeing global structure and local details. Besides, our method can also generate high-quality 3D humans guided by text. Experimental results demonstrate that our method outperforms the state-of-the-art methods, making it ideal for advanced 3D applications.

\noindent \textbf{Limitations.} Although our method can produce results with various poses, supporting extreme poses is still a huge challenge, such as stooping down or standing on the head. More failure cases are shown in the supplementary material.

\noindent \textbf{Broader Impact.} Our method will promote the development of avatar generation, which is useful for VR/AR applications and makes up for the lack of 3D human datasets. However, this may also cause privacy and ethical problems. We suggest policymakers establish an efficient regulatory system and inform users about potential risks.

\noindent \textbf{Acknowledgements.} 
This work was supported in part by National Key R$\&$D Program of China (2023YFC3082100), National Natural Science Foundation of China (62122058 and 62171317), and Science Fund for Distinguished Young Scholars of Tianjin (No. 22JCJQJC00040).

{
    \small
    \bibliographystyle{ieeenat_fullname}
    \bibliography{main}
}


\end{document}